# Gait Cycle-Inspired Learning Strategy for Continuous Prediction of Knee Joint Trajectory from sEMG

Xueming Fu[1], Hao Zheng[2], Luyan Liu[3], Wenjuan Zhong[1], Haowen Liu[1], Wenxuan Xiong[1],
Yuyang Zhang[1], Yifeng Chen[1], *Member, IEEE*, Dong Wei[2], Mingjie Dong[4],
Yefeng Zheng[2], *Fellow, IEEE* and Mingming Zhang[1] , *Senior Member, IEEE*

*Abstract*—Predicting lower limb motion intent is vital for controlling exoskeleton robots and prosthetic limbs. Surface electromyography (sEMG) attracts increasing attention in recent years as it enables ahead-of-time prediction of motion intentions before actual movement. However, the estimation performance of human joint trajectory remains a challenging problem due to the inter- and intra-subject variations. The former is related to physiological differences (such as height and weight) and preferred walking patterns of individuals, while the latter is mainly caused by irregular and gait-irrelevant muscle activity. This paper proposes a model integrating two gait cycle-inspired learning strategies to mitigate the challenge for predicting human knee joint trajectory. The first strategy is to decouple knee joint angles into motion patterns and amplitudes—former exhibit low variability while latter show high variability among individuals. By learning through separate network entities, the model manages to capture both the common and personalized gait features. In the second, muscle principal activation masks are extracted from gait cycles in a prolonged walk. These masks are used to filter out components unrelated to walking from raw sEMG and provide auxiliary guidance to capture more gait-related features. Experimental results indicate that our model could predict knee angles with the average root mean square error (RMSE) of 3.03±0.49 degrees and 50ms ahead of time. To our knowledge this is the best performance in relevant literatures that has been reported, with reduced RMSE by at least 9.5%.

*Index Terms*—Surface electromyography, gait cycle, continuous joint movement prediction, deep learning

*This work has been submitted to the IEEE for possible publication. Copyright may be transferred without notice, after which this version may no longer be accessible. This work was supported in part by the National Natural Science Foundation of China under Grant 62273173. (Corresponding authors: Mingming Zhang and Yefeng Zheng)

This study incorporated human participants as part of its research. The Ethics Committee of the Southern University of Science and Technology, Shenzhen, China, granted approval for all ethical and experimental procedures and protocols under Application No. 20190004.

[1]X. Fu, W. Zhong, H. Liu, W. Xiong, Y. Zhang, Y. Chen and M. Zhang are with the Shenzhen Key Laboratory of Brain-robot Rehabilitation Technology Lab, Department of Biomedical Engineering, College of Engineering, Southern University of Science and Technology, Shenzhen 518055, China (e-mail: zhangmm@sustech.edu.cn).

[2]H. Zheng, D. Wei and Y. Zheng are with Tencent Jarvis Lab, Shenzhen, China (e-mail: yefengzheng@tencent.com).

[3]L. Liu is with the Medical Image Computing Lab, School of Biomedical Engineering, ShanghaiTech University, Shanghai 201210, China.

[4]M. Dong is with the Beijing Key Laboratory of Advanced Manufacturing Technology, Faculty of Materials and Manufacturing, Beijing University of Technology, Beijing, 100124, P.R. China.).

## I. INTRODUCTION

WEARABLE lower-limb exoskeleton holds great promise in walking assistance [1]–[3]. To provide proper assistance, it is necessary to convey human movement intentions to the exoskeleton. This has been commonly achieved by decoding the signals from wearable sensors [4], [5] such as inertial measurement units (IMU) [6]–[8], gyroscopes [9], [10] and surface electromyography (sEMG) sensors [11]–[14]. Compared with the first two types of signals, sEMG signals, generated before the actual muscle activity, have the potential for a natural human-robot interface [15].

In recent years, machine-learning-based sEMG processing methods have been adopted in various tasks such as classification of hand gestures and gait events [16]–[19], continuous prediction of joint angle [11], [12] and muscle force [20], [21]. In lower-limb joint angle prediction, Yi et al. [22] built the map between the sEMG as well as IMU data and knee joint angles based on a sequential data processing module (Long Short-Term Memory) [23] and obtained an average RMSE of 3.98±0.69 degrees; Wang et al. [24] adopted convolutional and recurrent neural networks to decode sEMG signals for classifying gait phases and estimating continuous joint angles with RMSE being 3.75±1.52 degrees; Wang et al. [25] proposed a temporal convolutional network to predict lower-limb joint angles from sEMG and vibroarthrography data, and achieved an averaged RMSE of 3.43±0.42 degrees. Notably, these studies attempted to build direct mapping between raw sEMG and human joint angles, and achieved decent results via various neural networks. However, they ignored the inter-subject variations, which caused models to simply memorize training data rather than learning intrinsic associations between sEMG and joint trajectory.

To cope with the challenge of inter-subject variations, transfer learning strategy has been widely used with improved performance. For example, Côté-Allard et al. [26] adopted adaptive batch normalization, a form of transfer learning, to address inter-subject variability, under the assumption that batch normalization statistics capture differences between subjects. The results showed that classification accuracy improved by 6.36%-11.31% with adaptive batch normalization (BN) layers compared to without; In the same task, Zou et al. [27] utilized domain adaptation strategy, training model on both the labeled data from the source subjects and unlabeled data from the target subject to align their feature distributions. Results showed that the model adapted with target data performed 11% better than model trained without target data. These methods aim at learning more general and robust features that are effectively adapted to a new user. Compared with recognizing fixed hand gestures, the regression of continuous knee angles is more complicated as the trajectory varies among subjects due to different height, weight, gaits, etc. Significant differences including frequency and amplitude can be observed in the joint angle curve across subjects. For this situation, when directly training deep neural networks with joint angle labels, the model tends to just remember training data, thus the generalization ability is restricted.

Considering that the joint angle trajectories contain both subject-specific information and common patterns feature, to mitigate inter-subject variation, the learning of joint angle regression is divided into movement patterns estimation and amplitude prediction. Our method was inspired by the work of Gerald et al. [28] and adapted for joint trajectory decoding. They argued that previous methods directly embedded information from different dimensions of time series data into a feature vector. These models struggled to learn meaningful representations, and they were opaque easy to overfit. To overcome this problem, based on decomposition learning, they disentangled time series observations into seasonal and trend terms to separately represent periodic patterns and specific trend of the time series data. This decoupling architecture inspires the strategy for inter-subject differences problem. The elimination of amplitude differences helps the network to learn more robust features to capture walking patterns across subjects.

In addition to inter-subject difference, the intra-subject variation [29] also impacts the decoding of sEMG signals. It is well known that human lower limb muscles exhibit a large amount of redundancy, resulting in gait-irrelevant muscle activity that may deteriorate the performance of gait predictions. A relevant topic is the analysis of muscle activation patterns [30], [31], which aims at extracting the principal activation to assess the normal and pathological functions of human walking. Due to the stride-to-stride variability, muscle activation intervals are acquired from numerous gait cycles to eliminate the impacts of motion-irrelevant muscle activities. Inspired by these works, we adopted muscle activation as an important cue to resolve the intra-subject variation problem. The muscle activation is used as a filter upon the sEMG signals for the first time, and the prediction of activated intervals further guides the network to learn more gait-related features.

In the current study, we adopt a spatio-temporal convolutional network that integrates gait kinematic decoupling strategies and principal muscle activation strategies to predict knee joint trajectories during walking from sEMG. The contributions of this paper can be summarized as follows:

1) Gait kinematic decoupling is proposed to cope with inter-subject variation. The movement pattern and gait amplitude information are separately decoded through a multi-task learning framework.

2) Muscle principal activation is adopted to guide the learning of gait-relevant features to resolve the intra-subject variability problem.

3) Extensive experiment validated the improved performance of the proposed strategies for predicting human knee kinematics from sEMG.

## II. METHOD

This section presents the network model incorporating two gait cycle-inspired strategies, as illustrated in Fig. 1. The first strategy for inter-subject variation problem is the gait kinematic decoupling, corresponding to the joint angle prediction module in Step 2 in Fig. 1. The other strategy for intra-subject variation problem is the muscle-activation-pattern filtering strategy, involving timing predictor in Step 1 as well as activation filter and activation predictor in Step 2. In the joint angle predictor module, the motion patterns and amplitudes information are estimated by separate prediction heads, combining to generate the final joint angle. Before being fed into the joint angle predictor, the raw sEMG inputs are weighted by a muscle activation interval. To align the inputs and corresponding muscle activation intervals, the gait timing segments of the input are predicted by the timing predictor. To extract more gait-related features, the muscle activation prediction module is integrated to predict the corresponding muscle activation. In Step 3, the online test and demo are presented.

### A. Gait Kinematic Decoupling

First of all, the definitions and notations are introduced. Let $x(t)$ denote the sEMG vector sequence. And the knee angle sequence labels are represented as follows

$$y(t) = \mu(t) \cdot G(t) + \sigma(t), \quad (1)$$

where $y(t)$ is the joint angle at time $t$, $G(t) \in [-1,1]$ is the motion patterns of trajectory, $\mu(t)$ denotes the amplitudes of trajectory and $\sigma(t)$ is the offset. Let $y(t) = y_p(t) \circ y_a(t)$, where labels $y_p(t)$ denote gait patterns $G(t)$ and labels $y_a(t)$ denote gait amplitudes information $\mu(t)$ and $\sigma(t)$.

Fig. 2 illustrates the joint angle curves of three human subjects and their decoupled components. It is seen that significant cross-subject variations appear among their joint angle trajectories (Fig. 2-a). Based on the prior knowledge that the periodic generation of sEMG signals by muscles leads to continuous joint motion, we can deduce that $y_p(t)$ describes cyclic subject-invariant motion patterns information (Fig. 2-b) and $y_a(t)$ represents subject-independent motion amplitudes information (Fig. 2-d). Notably, $y_p(t)$ is primarily determined by the features of input sEMG signals and $y_a(t)$ is mainly determined by the physiological traits including height and weight as well as personal walking habits. When the joint angles are directly used as labels, the model tends to remember the mappings from sEMG signals to knee angles as both the signals and labels are clearly different among individuals. To alleviate the overfitting caused by this issue, we propose to separately learn two mappings, $f_p(\cdot): x(t) \to y_p$ and $f_a(\cdot): x(t) \to y_a$ to decode of gait motion patterns $y_p(t)$ and gait motion amplitudes $y_a(t)$.

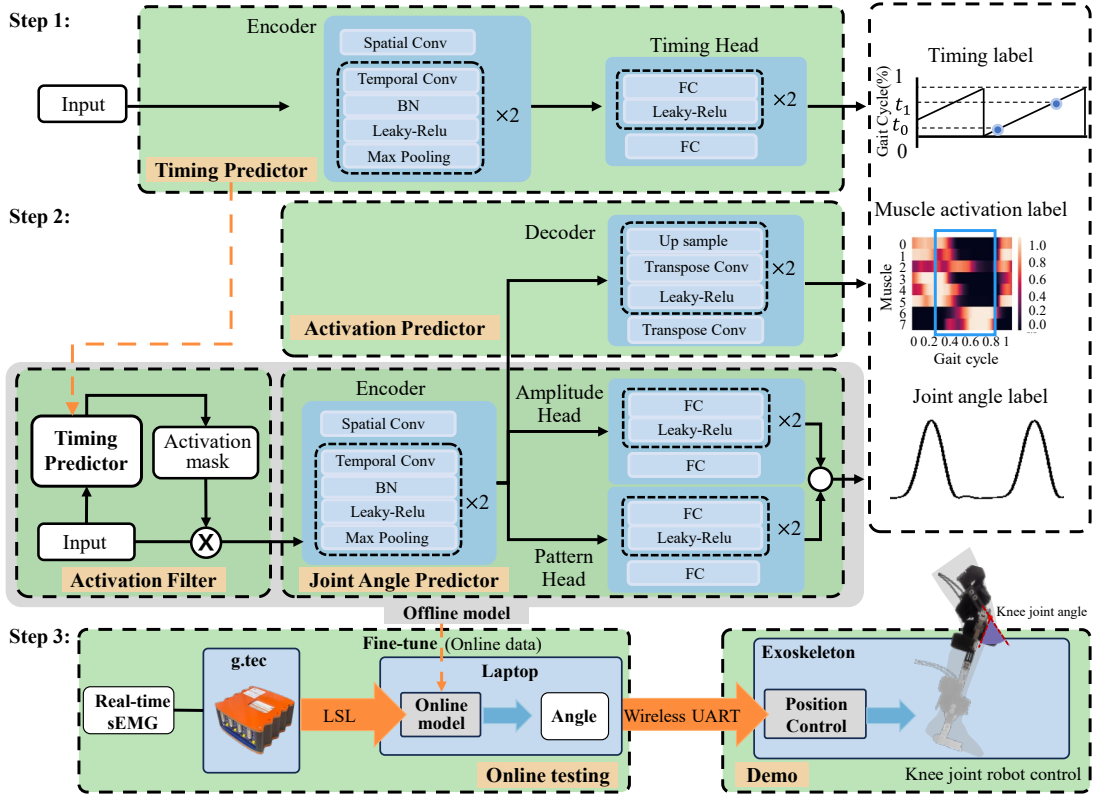

Fig. 1. The proposed joint angle prediction framework. It consists of three main steps: (1) training a timing predictor network to predict the gait timing of sEMG input; (2) utilizing the timing predictor to train an offline model, which includes a gait kinematics decoupling-based joint angle predictor module, a muscle activation filter module, and a muscle activation predictor module; (3) fine-tune the offline model for online testing and knee joint exosuit robot control demo. Note, "BN": Batch Normalization Layer; "FC": Fully-Connected Layer; "LSL": Lab Streaming Layer.

This strategy is achieved by the stepwise label decoupling. For each gait cycle (from peak to peak), the gait amplitudes labels are calculated as follows,

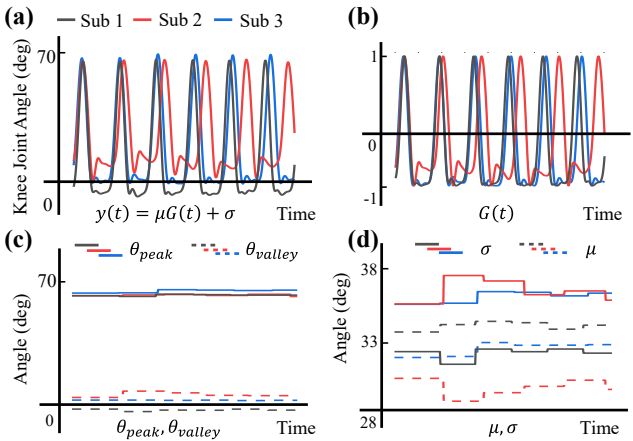

Fig. 2. Illustration of the gait kinematic decoupling. (a) The joint knee angle curves of three subjects. (b) The plots of decoupled gait pattern labels $G(t)$. (c) The series of maximums angle $\theta_{peak}$ and minimum angle $\theta_{valley}$. (d) The trajectory of amplitude factors $\mu$ and $\sigma$.

$$\mu = (\theta_{peak} - \theta_{valley})/2, \qquad (2)$$
$$\sigma = (\theta_{peak} + \theta_{valley})/2, \qquad (3)$$

where $\theta_{peak}$ and $\theta_{valley}$ denote the maximum and minimum joint angles in a gait cycle, respectively. Then the gait pattern labels are acquired by (4)

$$G(t) = (\theta(t) - \sigma)/\mu, \qquad (4)$$

In this way, the inter-subject variation becomes smaller in the gait patterns compared to the original joint angles. The prediction of amplitude information promotes the network to learn more robust features to capture the stride-to-stride and cross-subject amplitudes differences. The joint angles estimation network is composed of an encoder and two regression heads. The sEMG signals from eight muscle channels are truncated with a sliding window of 1s. With a sampling rate of 1200 Hz, the input consists of 1200×8 sample points, while the outputs including $G(t_1)$, $\mu_{t_1}$ and $\sigma_{t_1}$ only focus on the time after 50ms from the end time $t_1$ of this segment. A sliding window with a stride of 50ms is adopted. Mean square error (MSE) loss is adopted as the regression loss. The detailed structure of this network is illustrated in Table I. The encoder consists of three convolutional layers with the corresponding normalization and activation layers. Each regression head is composed of three fully-connected layers. The model is lightweight to accelerate the inference. The final predicted angle is acquired according to (1), (2) and (3).

TABLE I
DETAILED NETWORK OF THE FEATURE ENCODER, DECODER AND REGRESSION HEAD.

| Modules | Layers | Parameters | Output |
|---|---|---|---|
| Input | - | - | 1× 1200×8 |
| Encoder | Spatial Conv | 1×8, 8 | 8× 1200×1 |
|  | Permute D | - | 1× 1200× 8 |
|  | Temporal Conv | 10×1, 16 | 16× 1195 ×8 |
|  | Max Pooling | 5×1, stride 2 | 16× 239 ×8 |
|  | Leaky-Relu | 0.2 | 16×239 ×8 |
|  | BN | - | 16× 239 ×8 |
|  | Temporal Conv | 10×1, 16 | 16× 235 ×8 |
|  | Max Pooling | 5×1, stride 2 | 16× 47 ×8 |
|  | Leaky-Relu | 0.2 | 16× 47 ×8 |
|  | BN | - | 16× 47 ×8 |
| Decoder | Upsample | 5×1,"nearest" | 16× 235 ×8 |
|  | ConvTranspose | 5×1,16 | 16×239×8 |
|  | Leaky-Relu | 0.2 | 16× 239 ×8 |
|  | Upsample | 5×1, "nearest" | 16× 1195 ×8 |
|  | ConvTranspose | 6 ×1,1 | 1× 1200× 8 |
|  | Leaky-Relu | 0.2 | 1× 1200× 8 |
|  | Permutation | - | 8× 1200×1 |
|  | ConvTranspose | 1 × 8,1 | 1× 1200× 8 |
| Head | Fully-connected 1 | 6016,500 | 500 × 1 |
|  | Leaky_Relu | 0.2 | 500 × 1 |
|  | Fully-connected 2 | 500,50 | 50 × 1 |
|  | Leaky-Relu | 0.2 | 50 × 1 |
|  | Fully-connected 3 | 50,1(2) | 1×1(2×1) |

### B. Muscle-activation-pattern Filtering

Lower-limb sEMG signals contain numerous muscle activities, however, not all activities are related to gait walking, which is one of the major reasons for the stride-to-stride and intra-subject variability. The sEMG signals are separated into gait-related components $x_s(t)$ and gait-unrelated components $x_n(t)$. That is $x(t) = x_s(t) + x_n(t)$. The gait-related components can be represented by the muscle principal activation masks, which warp activation probability distributions across gait cycle. Thus, we hypothesized that the gait-related components $x_s(t)$ is approximately estimated by muscle activation $\hat{x}_s(t)$. As Fig. 3 illustrates, the muscle activation probability distributions of different muscles vary throughout a gait cycle. During the swing phase, the anterior thigh and shank muscles propel the limb forward (Fig. 3-C), causing large changes in knee joint angles (Fig. 3-A). Thus, these muscles exhibit relatively higher activation probabilities, as can be observed in Fig. 3-B. In stance phase, the posterior shank muscles provide limb support (Fig. 3-C). Therefore, they exhibit higher activation probabilities, as observed in Fig. 3-B.

In this study, we proposed two approaches to take advantage of the prior information—muscle activation masks. First, the muscle activation masks serve as filters for raw input. We propose to learn the map $f(\cdot): x(t) \times \frac{\hat{x}_s(t)}{||\hat{x}_s(t)||} \to y(t)$. The alignment of input segments to corresponding interval of masks is required. The extra timing predictor is proposed to predict the start time $t_0$ and the end time $t_1$ of input within the gait cycle, as illustrated in Step 1 of Fig. 1. The corresponding interval is cut from the average mask according to $t_0$ and $t_1$. Then, the intervals are interpolated in the same resolution as the input segments. The inputs are multiplied by intervals to filter out the gait-unrelated component. The detailed structure of the timing predictor is presented in Table I. The composition of the encoder is the same as feature extractor. The timing prediction head consists of three fully-connected layers, but the output channel number of the last layer is two.

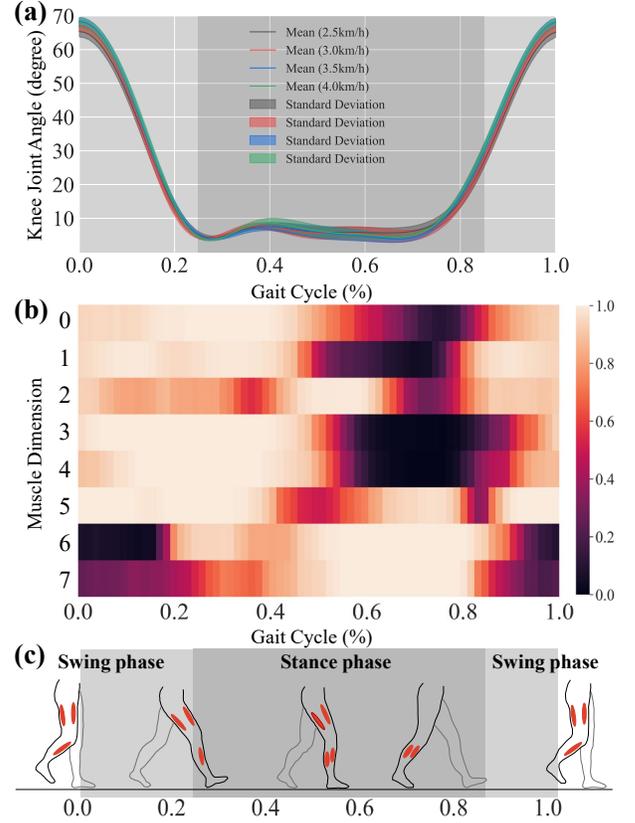

Fig. 3. Schematic diagram of muscle activation pattern corresponding to the gait cycle of one subject. (a) The average curves of knee joint trajectory under different speeds. (b) The average muscle activation probability heatmap. the vertical dimension axis from 0 to 7 represents eight muscles. (c) The schematic representation of the corresponding gait phase. The red ellipses indicate the primary locations of muscle activations. The dark shading indicates the support phase while the light shading indicates the swing phase.

In addition to a filter for the raw sEMG signals, the principal muscle activation is also used as a guide for feature extraction. This is achieved by predicting muscle activation as an auxiliary task. As shown in Fig. 1, the estimation of muscle activation and the prediction of joint angles share the same encoder, while a separate decoder is added to recover the principal muscle activation from encoded representations. This task imposes the shared encoder to capture more gait-related features. During testing, this branch is directly removed without extra computation cost. The detailed structure of the decoder is illustrated in Table I. Up-sampling and transposed convolutional layers are adopted to recover the original size. This part is also trained with the MSE loss.

To acquire these masks, we adopt a double-threshold detector [30] to extract the binary muscle activation from sEMG. The detector consists of four parameters: 1) the detection window length $l$, 2) the first activation threshold value $\varepsilon$, 3) the second minimum activation sample number threshold $r$, and the state minimum duration $n$. The minimum time unit is 5ms indicating that $l$ is not bigger than six. The states of every sample point in an observation window keep the same. The $\varepsilon$ is half of the quantile value of each sEMG channel. We set $r$ to 1. The extraction pipeline is as follows. First, the initial binary activation matrixes are obtained by determining whether the number of samples on each sEMG channel exceeding $\varepsilon$ within detection window exceeds $r$. Second, the random noise of the initial activation masks is eliminated through postprocessing. The states lasting less than $30ms$ are set to the opposite state. Then, the binary activation mask sequences are split into multiple gaits according to gait cycle timing. Finally, the average mask of multiple gaits that are resized to 100×8 is used as the muscle principal activation mask.

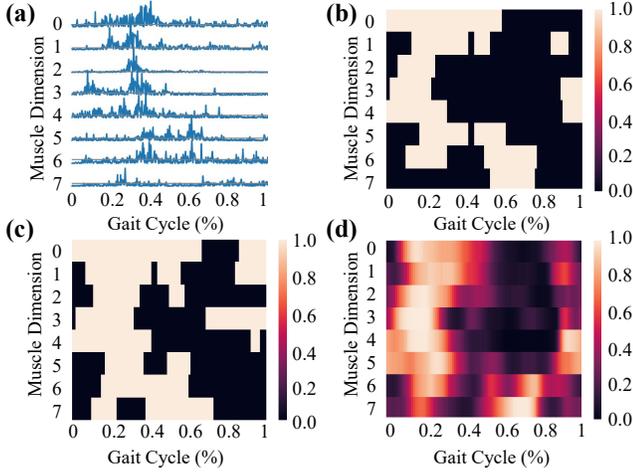

Fig. 4. Illustration of the muscle activation mask: (a) the raw sEMG signals of one stride; (b) the muscle activation mask corresponding to this step; (c) the muscle activation of another stride; (d) the principal activation of this subject obtained by averaging the muscle activation masks of enormous gaits.

Fig. 4 presents the activation masks of two different gaits from the same person. The outputs of the detector are binary masks in which the values for the non-active segment are zero. It is seen that the activation intervals of each muscle are different between steps. Since gait-irrelevant activity does not periodically occur at each gait cycle, the averaging operation acts as a filter for noise activation removal. Such masks are intrinsically used to suppress sEMG signals that are not related to walking.

### C. Network Training and Evaluation

As illustrated in Fig. 1, the overall process of model training includes three steps. Step 1, the timing predictor module based on the pre-processed sEMG is trained. Step 2, with the help of the timing predictor, the offline joint angle model is trained. Step 3, the offline model is fine-tuned for online testing and demonstration.

*1) Training Setting:*

The proposed model is optimized by adaptive moment estimation (Adam) with betas of (0.9, 0.99). The initial learning rate for the model is 0.0025. The learning rates are adjusted by cosine annealing with a cycle period of 20 epochs and a minimum value of $1\times 10^{-8}$. Early stopping is adopted to avoid overfitting. The model performance is evaluated by the mean value of all the participants. All the models are trained on a single NVIDIA Tesla V100 GPU with a batch size of 512.

In Step 1, prediction errors are inevitable for start and end timing. To mitigate the impacts of this error, we design a targeted data augmentation method, in which random perturbations are added to the start and end timing labels when cutting the muscle activation intervals. the standard MSE loss is adopted as the loss function for the timing predictor module,

$$L(\theta_f, \theta_{\text{timing}}) = \frac{1}{n}\sum_{i=1}^{n}(t_i^s - \hat{t}_i^s)^2 + \sum_{i=1}^{n}(t_i^e - \hat{t}_i^e)^2, \quad (5)$$

where $\theta_f$, and $\theta_{\text{timing}}$ are the parameters of the feature encoder and timing predicting head. $t_i^s$ and $\hat{t}_i^s$ are the label and prediction of the start gait timing of input, respectively. $t_i^e$ and $\hat{t}_i^e$ are the label and prediction of the end gait timing of input, respectively.

In Step 2, the main network is optimized via the following loss function,

$$L(\theta_f, \theta_{inf}, \theta_{amp}, \theta_p) = L_1 + L_2 + \alpha L_3, \quad (6)$$

$$L_1(\theta_f, \theta_{inf}) = \frac{1}{n}\sum_{i=1}^{n}(y_i^{inf_s} - \hat{y}_i^{inf_s})^2, \quad (7)$$

$$L_2(\theta_{amp}) = \frac{1}{2n}\sum_{i=1}^{n}\sum_{j=0}^{1}(y_{i,j}^{amp_s} - \hat{y}_{i,j}^{amp_s})^2, \quad (8)$$

$$L_3(\theta_f, \theta_p) = \frac{1}{800n}\sum_{i=1}^{n}\sum_{j=1}^{100}\sum_{k=1}^{8}(m_{i,j,k} - \hat{m}_{i,j,k})^2, \quad (9)$$

where $\theta_f$, $\theta_{inf}$, $\theta_{amp}$ and $\theta_p$ are the parameters of feature encoder, motion pattern head, amplitude head and activation predictor, respectively. $y_i^{inf}$ and $\hat{y}_i^{inf}$ are the pattern label and prediction, respectively; $y_{i,j}^{amp}$ and $\hat{y}_{i,j}^{amp}$ are the amplitude label and prediction, respectively. $m_{i,j,k}$ and $\hat{m}_{i,j,k}$ are the activation label and prediction, respectively.

In Step 3, the learning rate is 0.0005, and the number of training epochs is 10.

*2) Evaluation Metrics:*

To evaluate the performance of the prediction model, three evaluation metrics are adopted to quantify the accuracy of joint angle prediction, including the root mean square error (RMSE), the normalized root mean square error (NRMSE), and the coefficient of determination ($R^2$). The RMSE is as follows,

$$RMSE = \sqrt{\frac{\sum_{i=1}^{N}(\theta_i - \hat{\theta}_i)^2}{N}}, \quad (10)$$

where $\theta_i$ and $\hat{\theta}_i$ are the label and prediction of the knee joint angle, respectively.

The NRMSE formulas is as follows,

$$NRMSE = \frac{\sqrt{\sum_{i=1}^{N}\frac{(\theta_i - \hat{\theta}_i)^2}{N}}}{\theta_{max} - \theta_{min}}, \quad (11)$$

where $\theta_{max}$ and $\theta_{min}$ are the maximum and minimum joint angles, respectively.

The $R^2$ can be calculated as follows,

$$R^2 = 1 - \frac{\sum_{i=1}^{N}(\theta_i - \hat{\theta}_i)^2}{\sum_{i=1}^{N}(\theta_i - \bar{\theta}_i)^2}, \quad (12)$$

where $R^2$ reflects the fit between the predicted joint angles and the labels.

## III. EXPERIMENTS

### A. Participants and Experimental Setup

Ten healthy subjects (seven males and three females, age: 23 ± 4 years, height: 169 ± 17 cm, weight: 69 ± 23 kg) participated in the experiments for offline datasets. The study was approved by the Human Participants Ethics Committee of the Southern University of Science and Technology (20190004).

The experimental setup is illustrated in Fig. 5. Subjects were required to naturally walk on the treadmill at a stable speed. During the preparation stage, the subjects warmed up on the treadmill. When the treadmill reached the designated speed, the sEMG signals and optical motion capture data were recorded. The experiment consisted of four sessions with different speeds (2.5 km/h, 3 km/h, 3.5 km/h and 4 km/h). Each session comprised four trials, and each trial lasted at least three minutes. There was 30 second rest between trials and one minute rest between sections.

The online data communication environment from the sEMG collecting device to the exoskeleton needs to be established for online test and demo. Specifically, the Python interface to the Lab Streaming Layer (LSL) [33] is utilized to exchange the real-time time series data between a laptop and the LSL. The g.Needaccess client is used to stream real-time sEMG signals from the g.tec device to the LSL. For online demo, the joint angles are transmitted to the exoskeleton via the wireless universal asynchronous receiver/transmitter. The exoskeleton actuates the knee joint to the corresponding angles through position control. One subject participated in online test.

### B. Data Acquisition

Eight muscles of the right lower limb related to walking including rectus femoris (RF), vastus lateralis muscle (VL), vastus medialis muscle (VM), tibialis anterior muscle (TA), biceps femoris muscle (BF), semitendinosus (ST), gastrocnemius muscle media head (GM), and gastrocnemius muscle lateral head (GL) were chosen for sEMG signals. The sEMG signals of eight channels were recorded at a sampling rate of 1200 Hz by the device (g.tec medical engineering GmbH, Austria). Bipolar active electrodes were placed on the muscle at 20 mm intervals. Meanwhile, spatial position data of 15 markers on the body were collected at a sampling rate of 60 Hz by an eight-camera optical motion capture system (Raptor-4S, Motion Analysis Corporation, USA) in accordance with the Helen Hayes marker set [32].

### C. Data Preprocessing

The raw sEMG signals are preprocessed before being fed into the model as follows. The absolute value of raw sEMG signals passes through a low-pass 20 Hz and high-pass 500 Hz eighth-order Butterworth filter in turn. Then, they are processed by a 50 Hz notch filter. Finally, the filtered sEMG signals are normalized. Due to slight occlusions, the marker data of key points contained errors. After marker correction and filtering in Cortex software, smooth markers are obtained. Then, the standard knee joint angles are calculated from marker data as the "gold standard". The joint angles and sEMG signals are aligned at the start. Without loss of generality, we predict the right knee joint angles.

The data annotation is shown in Fig. 6. The sEMG signals are segmented into a series of sliding windows as the input. In this study, the length of the sliding window is 1s, and the sliding stride is 50 ms. The size of input is 1200×8. The joint angles are predicted 50 ms ahead of time. Therefore, the corresponding label for each input segment is the angle 50 ms after the end timing.

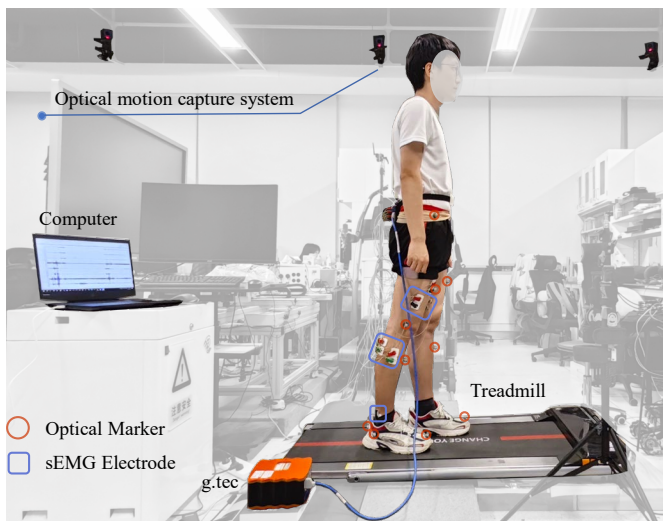

Fig. 5. Experimental setup

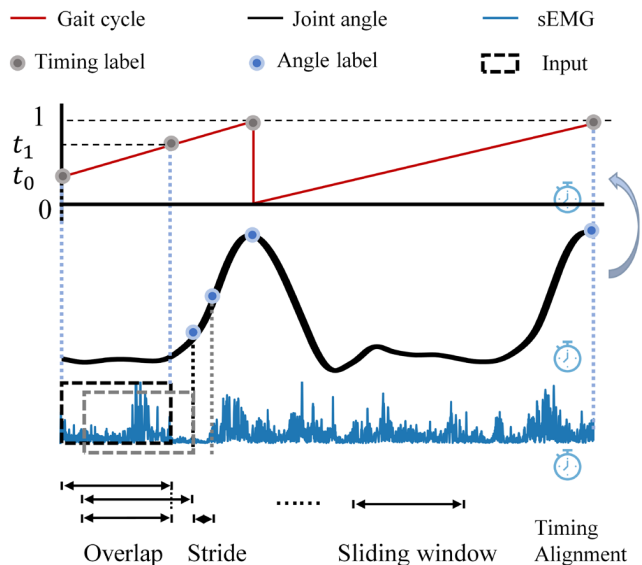

Fig. 6. Data annotation. The first row shows the gait cycle timing trajectory over time; The second row displays the joint angle trajectory over time; The third row presents the sEMG signal corresponding to kinematics information; The last row illustrates the sliding window setting used for segmenting the data.

## D. Offline and Online Test

The offline model was evaluated by the leave-one-subject-out principle. The 10-fold cross-validation is performed for ten subjects while 4-fold cross-validation corresponding to the four trials is conducted for one subject. For each fold, one trial of all speeds is selected for testing while the remaining data are for training. All comparison approaches followed the same setting.

For online test, a 20 second trial was conducted to obtain the mean and standard deviation of sEMG data for normalization. The data of a 10 second trial (3 km/h) was utilized for offline model finetuning. Another four 10 second trials under different speeds (2.5~4.0 km/h) were conducted to test the online performance. The online model also was used for simultaneous control of knee exoskeleton.

## IV. RESULTS AND DISCUSSION

### A. Offline Joint Angle Prediction Performance

To compare with the existing methods, we reproduced related models based on our datasets as shown in Table II. The numbers in the parentheses are the standard deviations. The detailed results for different subjects under various speeds are shown in Table III. TimesNet model was proposed by Wu et al. [33] to capture changes within and between time series periods and achieved an RMSE of 6.24 degrees. The temporal convolutional network model proposed by Bai et al. [34] was based on convolutional layers to model sequential data, and it achieved an RMSE of 4.23 degrees. Vaswani et al. [35] proposed the Transformer based on the attention mechanism, which performed excellently in decoding sequential data. The reproduced Transformer achieved an RMSE of 3.81 degrees. Yi et al. [22] decoded the sEMG by an LSTM model consisting of three LSTM layers and two fully-connected layers. This network achieved an RMSE of 3.65 degrees. Banville et al. [37] designed a lightweight spatio-temporal CNN model for sleep monitoring based on electroencephalography (EEG). This model was adapted for knee angle estimation and achieved an RMSE of 3.35 degrees. Due to the efficacy of the model, it was chosen as the baseline. Adaptive batch normalization (AdaBN) was adopted by Li et al. [36] to resolve the inter-subject variations in sEMG-based classification of hand gestures. However, the model with AdaBN layers performed poorer than the baseline without AdaBN layers in continuous prediction of knee joint angles. This transfer learning strategy for classification problems does not work for continuous prediction problems. The proposed model incorporating gait cycle-inspired learning strategies into the baseline achieved an RMSE of 3.03 degrees, which is a 9.6% improvement compared to the baseline. The results demonstrate the effectiveness of proposed strategies in continuous prediction of joint angles.

TABLE II
RESULTS OF THE KNEE JOINT ANGLE PREDICTION.

| Methods | RMSE (deg) | NRMSE | $R^2$ |
|---|---|---|---|
| Wu et al. [33] | 6.24 (1.14) | 0.087 (0.014) | 0.902 (0.038) |
| Bai et al. [34] | 4.23 (0.65) | 0.059 (0.009) | 0.954 (0.019) |
| Vaswani et al. [35] | 3.81 (0.53) | 0.053 (0.006) | 0.960 (0.014) |
| Yi et al. [22] | 3.65 (0.61) | 0.051 (0.008) | 0.965 (0.016) |
| Li et al. [36] | 3.48 (0.58) | 0.049 (0.008) | 0.966 (0.015) |
| Banville et al. [37] | 3.35 (0.45) | 0.048 (0.008) | 0.968 (0.015) |
| **Ours** | **3.03 (0.49)** | **0.043 (0.007)** | **0.976 (0.011)** |

Fig. 7 presents the knee joint angle predictions curves of baseline and our method at various speeds. The errors mainly arise in the regions around the peaks and valleys. In comparison to baseline, the curves of the proposed method are smoother and more stable in the vicinity of the extreme points. This indicates that the proposed method is more capable to capture the relationship between sEMG and the amplitude and bias characteristics of joint angle trajectory.

The performance of proposed model compared to the reported studies is summarized in Table IV. Given that in below knee prosthesis control applications, patients typically only have thigh muscles available, experimental results utilizing only five thigh muscles for continuous knee trajectory prediction are also presented in the table. The results show that our model achieves state-of-the-art performance. Previous works such as [24] and [25] utilized multiple sensor data to predict the knee joint angles, which perform better than using sEMG signals alone. However, our single sEMG approach that predicts 50 ms in advance achieves comparable and higher accuracy to these multi-sensor methods. Even with the utilization of only five thigh muscles, our approach achieves an average RMSE of 4.07 degrees. The result demonstrates the effectiveness of the proposed framework in extracting discriminative gait-related features from sEMG signals and the feasibility for knee joint prostheses control.

### B. Online Test Results

To shed more light on the potential issues of practical applications of sEMG-based joint angles prediction and provide references for other researchers, online testing and demo are carried out. The online average gait prediction curves under different speeds are shown in Fig. 8. The average gait curve predicted online aligns well with the online label curves. It also can be seen from the chart that online results show degraded performance. On the one hand, it may be caused by shifts of electrodes and markers across different experimental batches. For example, there are some differences between the label curves in the offline dataset (the second row and second column in Fig. 7) and the label plots in online data (Fig. 8). On the other hand, there is a long delay between the offline data collection time and online test time. The changes in the physiological state (e.g., thickness) of the lower limbs may cause variation in sEMG signals. The RMSE of online test is 4.38 degrees. It is seen that the errors mainly arise in the peaks.

Utilizing the computing resources of a laptop CPU, the inference time of proposed model is about 20 ms under offline test. To verify whether inference time and other latency meet the requirements for exoskeleton control applications, synchronized control demo of knee exoskeleton was performed. The model predicts angles 50 ms in advance. Fig. 9 shows the gait cycle curves of human and exoskeleton. The four representative frames correspond to four gait timings (0%, 17.8%, 60.3% and 72.2%). It can be observed that the total sum of online data pre-processing time, transmission time and exoskeleton latency fluctuated between 10 ms and 30 ms. However, the exoskeleton actuator acts prior to the human actual movement while the durations of different gaits intra subject are various. This demonstrates that the proposed method can meet the real-time requirements for online application.

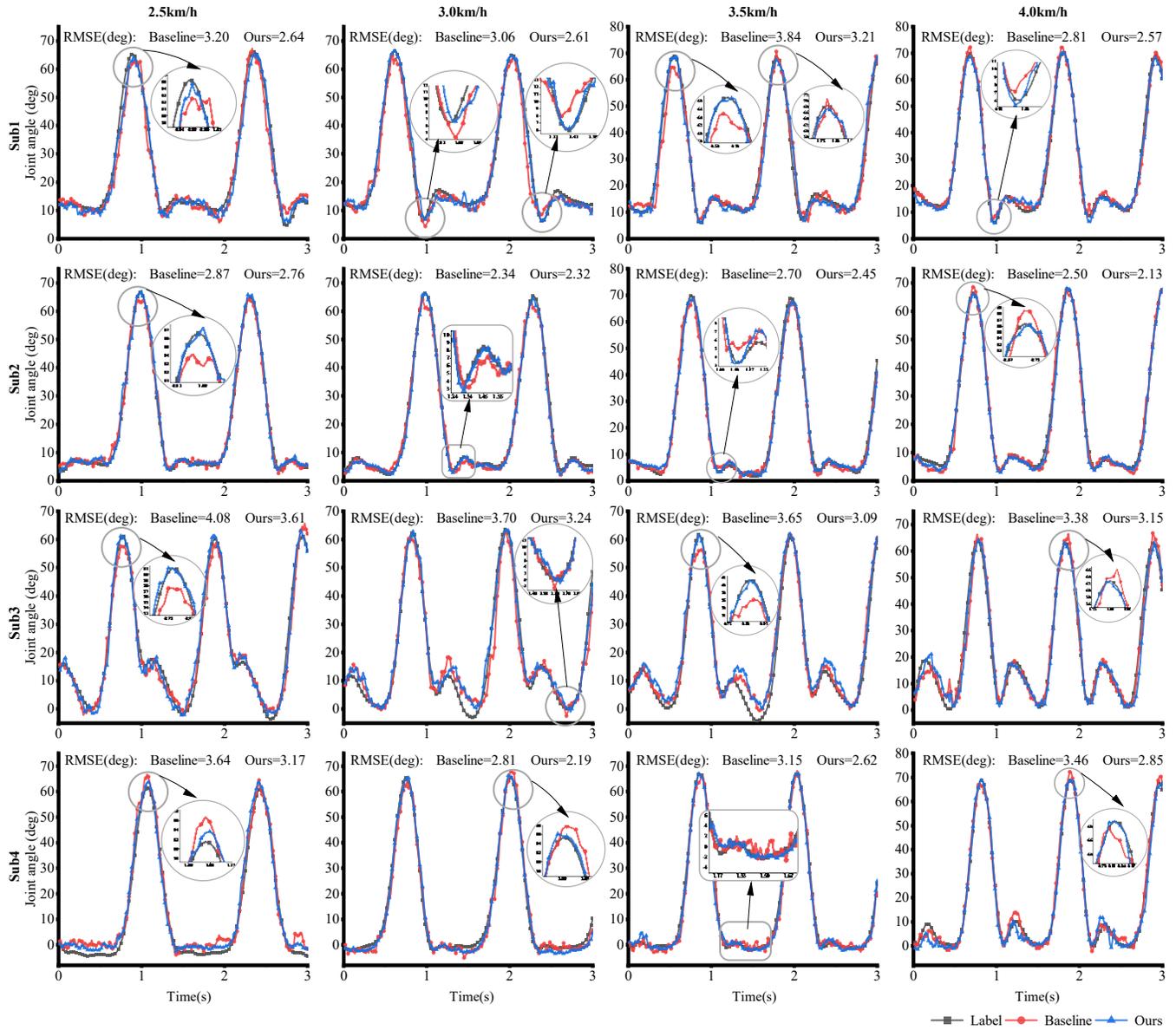

Fig. 7. Knee joint angle prediction curves of our proposed model versus the baseline model. Each row in order represents the results for the first four subjects, and each column represents the results under the walking speed of 2.5 km/h, 3.0 km/h, 3.5 km/h and 4.0 km/h, respectively.

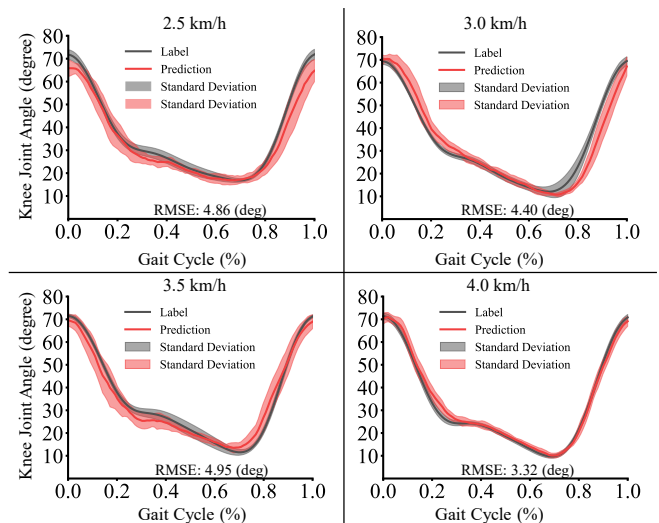

Fig. 8. Knee joint angle prediction curves of online tests.

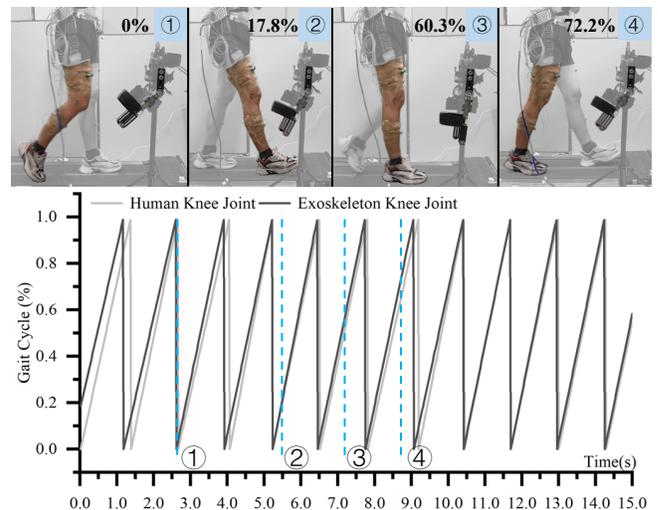

Fig. 9. Simultaneous control of knee exoskeleton during treadmill locomotion .

TABLE III
THE RESULTS OF THE KNEE JOINT ANGLE PREDICTION USING DIFFERENT METHODS UNDER DIFFERENT SPEEDS.

| Subject | Speed | [33] | [34] | [35] | [22] | [36] | [37] | Ours |
|---|---|---|---|---|---|---|---|---|
| Sub 1 | 2.5 km/h | 6.84 (0.74) | 5.04 (1.30) | 4.07 (0.62) | 2.84 (0.42) | 3.24 (0.12) | 3.20 (0.23) | **2.64 (0.10)** |
| | 3.0 km/h | 6.71 (0.42) | 3.89 (0.45) | 3.95 (0.11) | 2.77 (0.73) | 2.78 (0.09) | 3.06 (0.20) | **2.61 (0.18)** |
| | 3.5 km/h | 8.06 (0.43) | 5.25 (1.21) | 4.46 (0.12) | 3.88 (0.61) | 3.30 (0.08) | 3.84 (0.80) | **3.21 (0.71)** |
| | 4.0 km/h | 6.88 (1.05) | 3.83 (0.36) | 3.68 (0.45) | 3.33 (1.30) | 3.61 (1.00) | 2.81 (0.25) | **2.57 (0.20)** |
| Sub 2 | 2.5 km/h | 5.11 (0.34) | 3.47 (0.17) | 3.36 (0.31) | 3.01 (0.48) | 3.05 (0.38) | 2.87 (0.53) | **2.76 (0.41)** |
| | 3.0 km/h | 4.32 (0.34) | 2.80 (0.29) | 2.71 (0.15) | **2.31 (0.27)** | 2.47 (0.18) | 2.34 (0.11) | 2.32 (0.21) |
| | 3.5 km/h | 4.45 (0.17) | 3.03 (0.20) | 2.84 (0.17) | 2.74 (0.50) | 2.53 (0.07) | 2.70 (0.20) | **2.45 (0.10)** |
| | 4.0 km/h | 4.48 (0.50) | 3.20 (0.38) | 2.75 (0.25) | 2.33 (0.31) | 2.50 (0.16) | 2.50 (0.30) | **2.13 (0.18)** |
| Sub 3 | 2.5 km/h | 7.15 (0.51) | 4.73 (0.33) | 4.13 (0.28) | 4.48 (0.76) | 4.07 (0.36) | 4.08 (0.51) | **3.61 (0.27)** |
| | 3.0 km/h | 6.52 (0.41) | 4.25 (0.41) | 4.03 (0.13) | 3.98 (0.38) | 3.76 (0.22) | 3.70 (0.16) | **3.24 (0.11)** |
| | 3.5 km/h | 6.00 (0.80) | 4.07 (0.24) | 3.90 (0.33) | 4.04 (0.40) | 3.85 (0.73) | 3.65 (0.39) | **3.09 (0.31)** |
| | 4.0 km/h | 6.61 (0.34) | 4.44 (0.06) | 4.01 (0.20) | 3.90 (0.24) | 3.49 (0.11) | 3.38 (0.16) | **3.15 (0.14)** |
| Sub 4 | 2.5 km/h | 6.76 (1.15) | 4.08 (0.56) | 3.73 (0.19) | 3.90 (0.84) | 3.81 (1.06) | 3.64 (0.64) | **3.17 (0.63)** |
| | 3.0 km/h | 5.66 (0.15) | 3.52 (0.16) | 2.92 (0.16) | 2.76 (0.58) | 2.54 (0.17) | 2.81 (0.35) | **2.19 (0.23)** |
| | 3.5 km/h | 5.99 (0.52) | 3.97 (0.53) | 3.53 (0.37) | 2.79 (0.44) | 2.86 (0.20) | 3.15 (0.26) | **2.62 (0.27)** |
| | 4.0 km/h | 6.80 (1.40) | 4.44 (0.31) | 4.26 (0.34) | 4.27 (0.66) | 3.22 (0.26) | 3.46 (0.11) | **2.85 (0.22)** |
| Sub 5 | 2.5 km/h | 8.42 (1.53) | 4.38 (0.61) | 3.52 (0.38) | 3.54 (0.67) | 3.44 (0.21) | 3.68 (0.43) | **3.07 (0.45)** |
| | 3.0 km/h | 8.24 (0.47) | 3.89 (0.24) | 4.16 (0.49) | 3.52 (0.41) | 3.59 (0.34) | 3.70 (0.21) | **3.05 (0.28)** |
| | 3.5 km/h | 8.07 (0.37) | 3.86 (0.40) | 3.90 (0.44) | 3.51 (0.40) | 3.68 (0.35) | 3.74 (0.70) | **3.34 (0.59)** |
| | 4.0 km/h | 8.36 (0.61) | 4.59 (0.74) | 5.17 (2.52) | 4.03 (0.47) | 4.09 (1.41) | 3.87 (0.72) | **3.57 (1.34)** |
| Sub 6 | 2.5 km/h | 5.49 (0.53) | 3.60 (0.66) | 2.96 (0.20) | 3.28 (1.23) | 2.61 (0.13) | 2.84 (0.69) | **2.40 (0.45)** |
| | 3.0 km/h | 4.99 (0.60) | 3.07 (0.76) | 2.87 (0.24) | **2.46 (0.61)** | 2.58 (0.19) | 2.85 (0.56) | 2.53 (0.51) |
| | 3.5 km/h | 4.30 (0.22) | 2.79 (0.20) | 2.73 (0.17) | 2.33 (0.33) | 2.64 (0.28) | 2.47 (0.35) | **2.20 (0.44)** |
| | 4.0 km/h | 4.88 (0.40) | 2.85 (0.13) | 2.75 (0.37) | 2.41 (0.19) | 2.44 (0.20) | 2.53 (0.16) | **2.22 (0.19)** |
| Sub 7 | 2.5 km/h | 8.30 (1.88) | 6.09 (1.57) | 4.47 (0.87) | 4.79 (0.92) | 5.00 (0.92) | **3.97 (0.86)** | 4.02 (0.84) |
| | 3.0 km/h | 7.33 (0.51) | 5.19 (0.16) | 4.34 (0.52) | 4.00 (0.37) | 3.89 (0.26) | **3.38 (0.09)** | 3.85 (0.21) |
| | 3.5 km/h | 5.98 (0.81) | 4.89 (0.51) | 4.08 (0.40) | 4.24 (0.72) | 3.54 (0.44) | **3.43 (0.60)** | 3.50 (0.43) |
| | 4.0 km/h | 6.19 (0.35) | 5.04 (0.15) | 4.21 (0.08) | 3.90 (0.44) | 4.03 (0.24) | **3.29 (0.26)** | 3.53 (0.20) |
| Sub 8 | 2.5 km/h | 7.24 (0.97) | 5.28 (1.50) | 4.80 (1.71) | 5.34 (1.45) | 5.15 (1.74) | 4.90 (1.46) | **3.49 (1.82)** |
| | 3.0 km/h | 5.92 (0.16) | 4.35 (0.18) | 3.85 (0.23) | 4.08 (0.45) | 3.73 (0.08) | 3.69 (0.18) | **3.40 (0.23)** |
| | 3.5 km/h | 5.97 (0.33) | 4.39 (0.13) | 3.60 (0.24) | 3.54 (0.26) | 3.56 (0.13) | 3.33 (0.15) | **3.16 (0.24)** |
| | 4.0 km/h | 6.68 (1.38) | 4.54 (0.85) | **3.49 (0.19)** | 4.24 (1.31) | 3.72 (0.43) | 3.59 (0.53) | 3.55 (0.38) |
| Sub 9 | 2.5 km/h | 5.54 (0.32) | 3.92 (0.15) | 3.59 (0.29) | 3.62 (0.11) | 3.45 (0.31) | **3.26 (0.44)** | 3.30 (0.40) |
| | 3.0 km/h | 5.59 (0.11) | 3.84 (0.57) | 3.80 (0.29) | 3.28 (0.23) | 3.27 (0.29) | 3.18 (0.15) | **3.04 (0.24)** |
| | 3.5 km/h | 7.47 (2.69) | 6.69 (3.20) | 6.69 (2.27) | 7.23 (2.55) | 6.90 (2.58) | 5.91 (3.06) | **2.99 (0.10)** |
| | 4.0 km/h | 5.20 (0.42) | 3.64 (0.44) | 3.08 (0.13) | 2.93 (0.17) | 2.85 (0.14) | 2.66 (0.23) | **2.59 (0.14)** |
| Sub 10 | 2.5 km/h | 7.03 (1.49) | 5.57 (1.93) | 4.38 (0.33) | 5.20 (0.67) | **4.00 (0.42)** | 4.18 (0.75) | 4.00 (0.73) |
| | 3.0 km/h | 5.91 (0.45) | 4.55 (0.38) | 4.02 (0.13) | 4.23 (0.31) | 3.70 (0.30) | **3.51 (0.30)** | 3.52 (0.36) |
| | 3.5 km/h | 4.79 (0.32) | 3.86 (0.46) | 3.67 (0.18) | 3.31 (0.30) | 3.26 (0.33) | 3.30 (0.10) | **3.00 (0.14)** |
| | 4.0 km/h | 5.10 (0.49) | 4.06 (0.29) | 3.69 (0.34) | 3.35 (0.27) | 3.22 (0.41) | 3.22 (0.27) | **3.09 (0.29)** |
| Mean (std) | 2.5 km/h | 6.78 (1.12) | 4.61 (0.87) | 3.90 (0.56) | 4.00 (0.90) | 3.78 (0.81) | 3.66 (0.64) | **3.24 (0.54)** |
| | 3.0 km/h | 6.11 (1.13) | 3.93 (0.70) | 3.66 (0.59) | 3.34 (0.72) | 3.23 (0.57) | 3.22 (0.45) | **2.97 (0.54)** |
| | 3.5 km/h | 6.10 (1.38) | 4.28 (1.12) | 3.94 (1.09) | 3.76 (1.36) | 3.61 (1.23) | 3.55 (0.93) | **2.95 (0.40)** |
| | 4.0 km/h | 6.11 (1.18) | 4.06 (0.68) | 3.70 (0.75) | 3.47 (0.71) | 3.31 (0.58) | 3.13 (0.47) | **2.92 (0.53)** |

The result in bold indicates the best.

TABLE IV
COMPARISON WITH CURRENT RELATED RESEARCH. "SMG": SONO MYOGRAPHY; "VAG": VIBROARTHROGRAPHY; "+": DENOTES FUSING OTHER SENSORS.

| Research | Task | Data | Subjects | Method | Performance (knee joint) |
|---|---|---|---|---|---|
| Timothy et al. [8] | Selected speed | IMU | 12 subs | Hinge Model | RMSE=7.87±3.27 |
| Mohammad et al. [43] | - | Ultrasound imaging | 7 subs | Gaussian Process Regression | RMSE=7.45±3.72 |
| Alireza et al. [44] | - | sEMG (5 muscles) | 19 subs (175-186 cm) | LSTM | RMSE=6.77±1.20 |
| Xiong et al. [38] | 0.5, 1.0, 2.0, 3.0 km/h | sEMG (8 muscles) | 8 subs (170-182 cm) | BiGRU | RMSE=6.28±0.81 |
| Yi et al. [22] | Selected speed | IMU and sEMG (9 muscles) | 10 subs (170-180 cm) | LSTM | RMSE=3.98±3.26 (+IMU) |
| Chen et al. [39] | 2.9, 3.6, 4.3 km/h | sEMG (10 muscles) | 6 subs (166-180 cm) | Deep Belief Network | RMSE=3.96±0.69 |
| Wang et al. [24] | 3.0, 3.5, 4.0 km/h | sEMG (8 muscles) | 3 subs (170-175 cm) | Multi-branch Neural Network | RMSE=3.75±1.52 |
| Kaitlin et al. [40] | 2.8 km/h | SMG and sEMG (8 muscles) | 9 subs (161-183 cm) | Gaussian Process Regression | RMSE=7.24±1.80 (+SMG: 3.77±0.81) |
| Wang et al. [25] | 2.5, 3.0, 3.6 km/h | VAG and sEMG (8 muscles) | 10 subs (164-175 cm) | Temporal Convolutional Network | RMSE=4.41 ± 1.11 (+VAG: 3.43 ± 0.42) |
| **This work** | **2.5, 3.0, 3.5, 4.0 km/h** | **sEMG (8 muscles)** **sEMG (5 muscles)** | **10 subs (156-186 cm)** | **Our model** | **RMSE=3.03±0.49** **RMSE=4.07±0.58** |

The result in bold indicates our work.

## C. Ablation Study

To analyze the key components in the proposed framework, the spatio-temporal convolution structure [37] was chosen as the baseline and gradually added the gait kinematic decoupling and muscle-activation-based methods, respectively. The quantitative results are illustrated in Table V. After separating the joint angle regression into movement pattern estimation and amplitude prediction, the mean RMSE was from 3.35 to 3.23 degrees. More significant improvement is observed in the results of the muscle-activation-based filtering module, which achieved an RMSE of 3.19 degrees. The combination of gait kinematic decoupling and either of the muscle-activation-based methods can further improve the accuracy. However, the addition of activation prediction did not result in a boost in the precision of the filtering module. Finally, the best result is achieved by combining all these components.

The results of these methods under different amounts of target data are compared. Generally, the prediction errors in the training set reduce with the number of trials increasing. However, a clear gap is seen between the results with and without target data in the training set. Since the muscle principal activation masks needs the real angle trajectory for gait segmentation, the results of muscle-activation-related methods are unavailable in the "0 Trials" column. Compared with the baseline method, gait kinematic decoupling reduces the RMSE from 8.20 to 7.69 degrees. A decrease over 0.4 degrees is observed from one to three trials. The results are further improved after incorporating the muscle-activation-based filtering module. More specifically, after adopting the proposed framework, the RMSE in the result of two trials are considerably lower than the baseline result with three trials of target data, demonstrating effectiveness of our method.

The prediction performance of three different muscle activation thresholds is shown in Table VI. In comparison, using the mean muscle activation as the threshold achieves better performance than using the first quartile and third quartile as thresholds.

## D. Gait Kinematic Decoupling Module

Gait kinematic decoupling is proposed to alleviate the impacts of inter-subject variation. It also provides a new view to analyze the efficiency of the designed modules. As illustrated in Table VII, the RMSE is calculated for different components: movement pattern $G(\omega t)$, amplitude $\mu$ and offset $\sigma$. Compared with the results of the baseline, gait kinematic decoupling can improve the accuracy of each item. When the network is directly trained by the joint angle label, due to the inter-subject variations, the generalization capability of the learned features is limited. After decoupling, the normalized movement pattern curve is easier to learn. Besides, the prediction of amplitudes forces the model to extract related features under both inter- and intra-subject variations. As a result, the errors in $\mu$ and $\theta$ are also reduced. In contrast, muscle-activation-based modules only contribute to the estimation of movement patterns. However, although the errors of activation-based modules are larger than that of gait decoupling in each separate item, the overall RMSE is slightly lower. The reason is that the fluctuation at the extreme points becomes more significant without gait decoupling, as shown in Fig. 7, resulting in larger errors for $\mu$ and $\sigma$. Such errors can be compensated by more accurate predictions in other regions. Moreover, combining the muscle-activation-pattern filtering methods and gait decoupling can further reduce the errors in each component. This means that with suitable guidance, the filtering of gait-irrelevant signals also contributes to the estimation of step amplitude.

TABLE V
ABLATION EXPERIMENT ABOUT THREE KEY MODULES. "A", "B", "C" REFER TO THE GAIT KINEMATIC DECOUPLING, ACTIVATION PREDICTION, AND ACTIVATION FILTER FOR THE RAW SIGNALS, RESPECTIVELY. "TRIAL" DENOTES THE NUMBER OF TRIALS FROM THE TARGET SUBJECT IN THE TRAINING SET.

| A | B | C | 0 Trials | 1 Trial | 2 Trials | 3 Trials |
|---|---|---|---|---|---|---|
| × | × | × | 8.20 (2.43) | 3.75 (0.56) | 3.46 (0.52) | 3.35 (0.67) |
| √ | × | × | 7.69 (2.82) | 3.52 (0.47) | 3.29 (0.52) | 3.23 (0.54) |
| × | √ | × | - | 3.62 (0.54) | 3.31 (0.50) | 3.21 (0.49) |
| × | × | √ | - | 3.56 (0.53) | 3.28 (0.47) | 3.19 (0.50) |
| √ | √ | × | - | 3.51 (0.48) | 3.27 (0.52) | 3.17 (0.52) |
| √ | × | √ | - | 3.48 (0.49) | 3.24 (0.50) | 3.16 (0.53) |
| × | √ | √ | - | 3.54 (0.48) | 3.20 (0.46) | 3.14 (0.52) |
| √ | √ | √ | - | **3.44 (0.51)** | **3.14 (0.48)** | **3.03 (0.49)** |

The result is RMSE (deg).
The result in bold indicates the best.

TABLE VI
PREDICTION RESULTS OF DIFFERENT ACTIVATION THRESHOLD VALUES.

| Threshold | RMSE (deg) | NRMSE | $R^2$ |
|---|---|---|---|
| Q1 | 3.28 (0.89) | 0.046 (0.008) | 0.971 (0.014) |
| **Q2** | **3.03 (0.49)** | **0.043 (0.007)** | **0.976 (0.011)** |
| Q3 | 3.20 (0.87) | 0.040 (0.007) | 0.970 (0.015) |

"Q1", "Q2": and "Q3" denote the first quartile, mean, and third quartile, respectively. The result in bold indicates the best.

TABLE VII
QUANTITATIVE RESULTS OF THE PROPOSED METHODS IN ESTIMATING DIFFERENT COMPONENTS AFTER GAIT KINEMATIC DECOUPLING.

| Method | Angle | $G(\omega t)$ | $\mu$ | $\sigma$ |
|---|---|---|---|---|
| Baseline | 3.35 (0.67) | 0.109 (0.033) | 1.68 (0.41) | 2.00 (0.39) |
| +GKD | 3.23 (0.54) | 0.098 (0.023) | 1.34 (0.40) | 1.25 (0.35) |
| +MAF | 3.19 (0.50) | 0.170 (0.187) | 1.50 (0.28) | 1.92 (0.27) |
| **Proposed** | **3.03 (0.49)** | **0.095 (0.019)** | **1.23 (0.35)** | **1.21 (0.33)** |

"GKD": Gait Kinematic Decoupling module. "MAF": Muscle-activation-based Filtering module. The result in bold indicates the best.

## E. Muscle-activation-based Filtering Module

Surface EMG signals arise from the superposition of electrical signals generated by numerous motor neurons, which contain both motion-related and motion-unrelated ones. The muscle-activation-based filtering module is proposed to eliminate the motion-unrelated signals. The most important is activation mask, which represents muscle activation probability distribution during a gait cycle. In the time domain, the masks suppress the amplitudes of irregular muscle activity interval. Thus, the model assigns more attention to these intervals, where signals of interest are more likely to manifest.

DeepLIFT [44] analysis results are shown in Fig. 10. The sEMG heat map after the muscle-activation-based filtering highlights periods of meaningful muscle movement while filters out unnecessary sEMG signals. Then, the contribution score map indicates that the models are sensitive to sEMG signal intervals with high amplitudes. The signals of high amplitude are prone to "false positives"—misidentifying unnecessary muscle movements, tremors or other irregular activity as meaningful activation. Without proper filtration, the models could not distinguish gait-relevant signals from indifferent muscle activity. Models trained on filtered sEMG data concentrate on intervals of gait-relevant activation. Irrelevant noise and irregular muscle activity are suppressed, enabling the model to focus on purposeful signals contributing to natural locomotion. This demonstrates that muscle activation masks are effective for highlighting task-relevant sEMG. Models trained on these data seem to develop a more nuanced understanding of intentional muscle activity and dynamism, rather than responding broadly to the overall signal complexity. In summary, muscle activation masks provide a way to guide models to prioritize meaningful sEMG over irregular noise, enabling improved detection and classification of purposeful muscle activity.

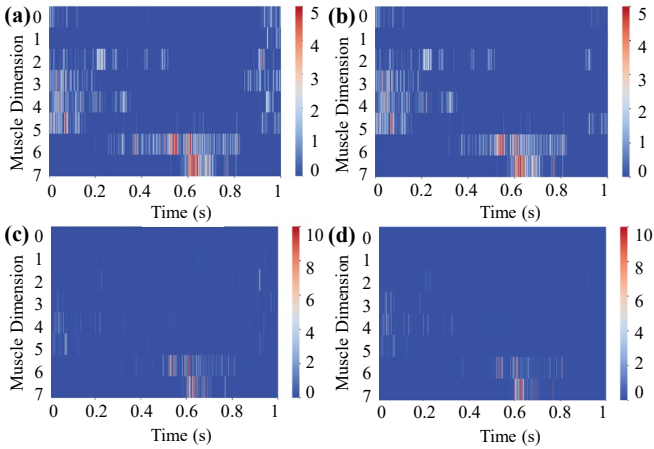

Fig. 10. Heatmap of sEMG signals, as well as contribution score: (a) sEMG signals; (b) sEMG signals after activation mask filtering; (c) contribution score; (d) contribution score after activation mask filtering.

Empirical modal decomposition (EMD) algorithms proposed by Huang et al. [45] is a data-driven technology to decompose non-stationary time series signals into different intrinsic model functions (IMFs) based on local characteristic time scales. Multi-empirical modal decomposition (MEMD) [46] is an improved version of EMD that can effectively tackle mode mixing issues. Low-frequency IMFs curves obtained by MEMD are found to be similar to the muscle activation masks curves obtained by the double-threshold detector. The antepenultimate intrinsic mode function obtained by MEMD has the longest period and most stable oscillations. The close match between the activation masks obtained by the double threshold approach and the average low-frequency-IMFs indicates that the double-threshold masks contain low-frequency trend item features in the multi-dimension sEMG. This hypothesis is verified in the following.

To gain deeper insights into the performance upper limits of the muscle activation filtering module and excavate the underlying principles behind it, additional offline experiments are implemented. The experiments compare the prediction results of two different mask-obtaining methods and directly using masks as inputs and weights. In these experiments, the masks and inputs are aligned by the accurate timing labels. The RMSE results are illustrated in Table VIII. The muscle activation mask obtained by the double-threshold detector and the average low-frequency IMFs masks are comparable. It indicates that the effective features contained in muscle activation pattern masks and low-frequency IMFs are similar. More interestingly, using these masks directly as input yielded better performance than using them as weights. This confirms that the human muscle activation patterns are highly relevant for predicting limb movements.

### F. Interpretation for Joint Angle Prediction

To gain insight into the proposed model, interpretability is analyzed using DeapLIFT. Fig. 11 illustrates the contributions of sEMG signals from different muscle channels. Specifically, the different muscles contribute differently to the model during different gait phases. During the stance phase, the biceps femoris (BF) muscle of the hamstrings make the greatest contribution. The BF muscle contributes a relatively high proportion throughout the gait cycle according to the model. In comparison, the model assigns more attention to the media head of the gastrocnemius muscle (GM) and the lateral head of the gastrocnemius muscle (GL) in the swing phase. The model captures this nuanced difference in the function of the GM and GL across gait phases. It assigns more attention to these muscles in stance versus swing, consistent with their higher electromyographic activity during support. During the swing

TABLE VIII
PERFORMANCE UPPER LIMITS OF MAF MODULE.

| Subjects | $DT_I$ | $DT_W$ | $MEMD_I$ | $MEMD_W$ |
|---|---|---|---|---|
| Sub 0 | 2.09 (0.11) | 2.15 (0.13) | 2.14 (0.13) | 2.08 (0.14) |
| Sub 1 | 1.79 (0.02) | 1.79 (0.06) | 1.76 (0.01) | 1.89 (0.22) |
| Sub 2 | 2.52 (0.21) | 2.54 (0.19) | 2.59 (0.13) | 2.50 (0.26) |
| Sub 3 | 2.61 (0.27) | 2.78 (0.57) | 2.26 (0.08) | 2.50 (0.17) |
| Sub 4 | 2.62 (0.17) | 2.69 (0.23) | 2.60 (0.16) | 2.60 (0.14) |
| Sub 5 | 1.82 (0.09) | 1.93 (0.14) | 1.80 (0.08) | 1.88 (0.23) |
| Sub 6 | 3.01 (0.09) | 4.40 (0.23) | 3.11 (0.11) | 4.49 (0.09) |
| Sub 7 | 3.06 (0.17) | 3.21 (0.25) | 2.94 (0.13) | 3.11 (0.18) |
| Sub 8 | 3.30 (0.65) | 2.91 (0.88) | 3.27 (0.61) | 2.93 (0.76) |
| Sub 9 | 2.83 (0.21) | 3.51 (0.39) | 2.81 (0.17) | 3.29 (0.35) |
| Mean (std) | 2.59 (0.52) | 2.79 (0.78) | **2.53 (0.52)** | 2.73 (0.79) |

"$DT_I$": double threshold masks used as inputs; "$DT_W$": double threshold masks used as weights; "$MEMD_I$": MEMD masks used as inputs; "$MEMD_W$": MEMD masks used as weights; The result in bold indicates the best.

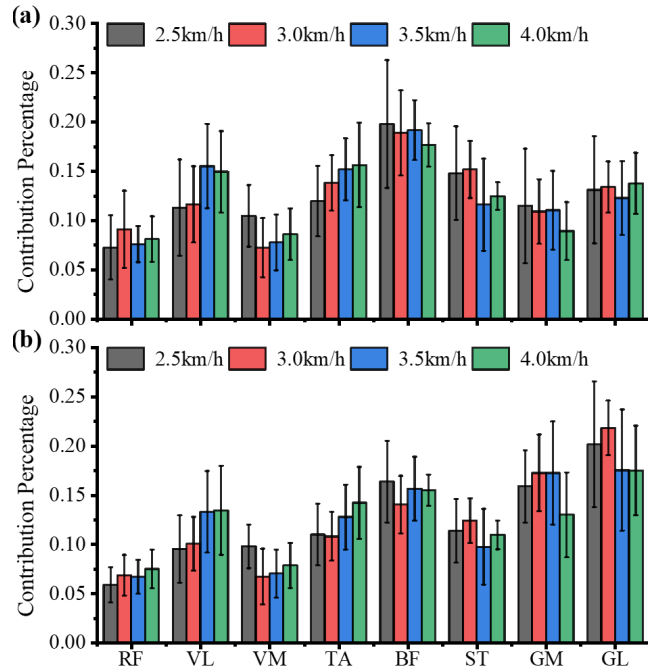

Fig. 11. Muscle contribution percentage via DeepLIFT: (a) prediction of joint angles during the stance phase; (b) prediction of joint angles during the swing phase.

phase, the model relies more on sEMG data from the previous stand phase in the historical input windows.

As depicted in Fig. 3, the activation of GM and GL muscles exhibit the greatest difference between the stance phase and the swing phase. To be specific, the GM and GL maintain a high level of activation during the stance phase while exhibiting a low-level activation during the swing phase. Since the sEMG signals from GM and GL among selected muscles have relatively higher activation during the stance phase, the model assigns more attention to these two muscles. This contrast highlights how these calf muscles contribute differently to knee extension and limb support compared with knee flexion and limb advancement during walking. In the stance phase, the GM and GL heads help stabilize the joint and transmit forces from the ground. But during the swing phase, these muscles play a less role, as the hip muscles drive the progression of the limb.

## V. Conclusion

In this paper, a new estimation framework incorporating two gait cycle-inspired learning strategies was proposed aiming at predicting human joint angles ahead-of-time actual motion from sEMG. The innovations consist of decoupled human motion information (to cope with the challenge of inter-subject variations) and muscle principal activation filter (to the challenge of intra-subject variations). Extensive experiments were conducted with human users, the effectiveness of the proposed joint prediction framework was verified in both offline and online environments. Considering the potential impact of human-robot interaction on sEMG signals, it is worth mentioning that the proposed framework may not be appropriate for direct robot control with human in the loop.